# Unsupervisedly Learned Representations – Should the Quest be Over?

Daniel N. Nissani (Nissensohn)

dnissani@post.bgu.ac.il

**Abstract.** After four decades of research there still exists a Classification accuracy gap of about 20% between our best Unsupervisedly Learned Representations methods and the accuracy rates achieved by intelligent animals. It thus may well be that we are looking in the wrong direction. A possible solution to this puzzle is presented. We demonstrate that Reinforcement Learning can learn representations which achieve the same accuracy as that of animals. Our main modest contribution lies in the observations that: a. when applied to a real world environment Reinforcement Learning does not require labels, and thus may be legitimately considered as Unsupervised Learning, and b. in contrast, when Reinforcement Learning is applied in a simulated environment it does inherently require labels and should thus be generally be considered as Supervised Learning. The corollary of these observations is that further search for Unsupervised Learning competitive paradigms which may be trained in simulated environments may be futile.

## 1. Introduction

This is neither a "new results" nor a "new method" paper, rather a "quo vadis" call to our Unsupervised Learning research community.

Corroborated by observations, it is widely believed that the animal world, humans included, learn rich world representations by some Unsupervised Learning scheme. Yet, in spite of at least four decades of research in this area, there still exists a Classification accuracy gap of about 20% between our best Unsupervised Learning methods and the corresponding skills exhibited by humans. We deduce this from the facts that a. humans perform very close to our state-of-the-art Supervised Learning methods (a.k.a. deep neural networks) in e.g. challenging Classification tasks (Russakovsky et al., 2015); and b. an accuracy gap of about 20% exists between these Supervised Learning methods and our best Unsupervisedly Learning schemes.

We provide next a brief survey of related work and published evidence for this second assertion above.

As hinted above, the search for Unsupervisedly Learned Representations has been strongly inspired by observation of the animal world[1] and further motivated by the high labeling costs of data sets of increasing size and complexity. The hope has been to be able to map sensorial input spaces into feature or representation spaces so that the mapped class conditioned distributions achieve better inter-class linear separability than that of the input sensorial space (linear separability has been the goal since the last layer of a neural net may be viewed as a linear classifier).

[1] See e.g. Yann LeCun Interview, IEEE Spectrum, Feb. 2015: "…The bottom line is that the brain is much better than our model at doing unsupervised learning…"

We will exclude from our survey clustering methods (like k-Means, spectral clustering and the like) since they operate on the input space itself and do not execute any input to feature space mapping; and will also exclude handcrafted or engineered so called feature extractors (e.g. Lowe, 1999) that while mapping an input to a representation space, their map is fixed and not trainable.

The vast majority of this effort has been focused in the area of Autoencoders, originally proposed by (Sanger, 1988): neural nets which attempt to encode an input into a more useful representation and then decode this representation back into the same input. In many cases the unsupervised training of such a network is executed by means of standard backpropagation techniques (Bishop, 2006) where the input reconstruction error is regularized by some intuitive criterion such as 'sparsifying' (Olshausen and Field, 1996; Ranzato et al., 2007), 'contractive' (Rifai et al., 2011), 'denoising' (Vincent et al., 2010; Bengio et al., 2011). In other cases reconstruction is implemented by stochastic methods (Hinton et al., 1995; Salakhutdinov and Hinton, 2007). Typically in these works several single layer Autoencoders are sequentially trained and layer-wise stacked. Such stacking can be viewed as an input space to representation space mapping, and could in principle have been evaluated in conjunction with a classifier layer (e.g. SVM, linear discriminant, etc.); no such evaluation has been apparently reported. Instead, after training, Autoencoder stacks have been generally used to provide parameters initialization of (supervisedly trained) deep neural nets. These have been shown to exhibit faster convergence vs. random weights initialization (Bengio et al., 2011; Bengio, 2012; Arora et al., 2014; Erhan et al., 2010). For a more comprehensive review the reader is referred to (Bengio and Courville, 2014).

Another, smaller but more recent and relevant corpus of works, use an auxiliary task in order to unsupervisedly train a deep neural net and achieve such a representation mapping. These include context prediction (Doersch et al., 2016; Pathak et al., 2016), puzzle solving (Noroozi and Favaro, 2016), video patches tracking (Wang and Gupta, 2015), split image prediction (Zhang et al., 2017), and Generative Adversarial Networks (Radford et al., 2016; Donahue et al., 2017). The Classification accuracies of these unsupervisedly learned representations (evaluated in conjunction with a classifier) are within a few percent from one another and about 20% worse than a fully supervisedly trained similar architecture net: e.g. 32.2% vs. 53.5% (Donahue et al., 2017) and 38.1% vs. 57.1% (Noroozi and Favaro, 2016), unsupervised vs. supervised respectively, all Top-1 on ImageNet dataset.

Since, as we have seen, Unsupervised Learning in the animal (explicitly – human) world performs similarly as well as our artificial Supervised Learning does, and this last outperforms our artificial Unsupervised Learning by about 20%, we conclude by transitivity that, in spite of decades of research, the accuracy gap between the two Unsupervised Learning schemes, the 'natural' and the 'artificial', remains about 20% large, as asserted above.

We may have been looking at the wrong direction.

As we will demonstrate in the sequel Reinforcement Learning (RL) methods can learn representations which when applied to a linear classifier achieve Classification accuracy competitive with that of similar architecture Supervised Learning neural net classifiers.

Furthermore and more importantly, simple gedanken-experiments will lead us to the conclusion that if a RL scheme is applied in *real-world environments* (e.g.in the animal world) then labels of no sort are involved in the process, and it thus may be legitimately considered an Unsupervised Learning scheme; in contrast, application of RL methods in the context of *simulated or symbolic processing environments* (as is usually the case both during research and in business applications) will necessarily require labels for learning and is thus doomed to remain, with some exceptions, a Supervised Learning scheme[2].

[2] So that, like Moses on Mount Nebo, we may be able to see Promised Land but will never reach it (Deuteronomy 34:4)

In Section 2 we will provide a concise introduction to RL concepts. In Section 3 we bring up a simple demonstration of representation learning by a RL method, which achieves competitive Classification accuracy relative to state of the art neural nets of similar architecture. We discuss our results and their implications in Section 4, and conclude in Section 5.

## 2. The Reinforcement Learning Framework in a Nutshell

RL has reached outstanding achievements during the last few decades, including learning to play at champion level games like Backgammon (Tesauro, 1994), highly publicized Chess and Go, and standing behind much of autonomous vehicles technology advances.

It is our goal in this brief Section to promote a connection, from Pattern Recognition community perspective, with the Reinforcement Learning methodology. There seems to exist, surprisingly, a relative disconnect that lasted for decades. For an excellent text on the RL subject we refer to (Sutton and Barto, 2018). Readers knowledgeable of RL may directly skip to the next Section.

The basic RL framework consists of an Agent and an Environment. The Agent observes the Environment States (sometimes a.k.a. Observations), and executes Actions according to some Policy. These Actions, in general, affect the Environment, modifying its State, and generating a Reward (which can also be null or negative). The new State and Reward are observed by the Agent and this in turn generates a new Action. The loop may go on forever, or till the end of the current episode, which may then be restarted. Actions may have non-immediate effects and Rewards may be delayed. A possible goal of the Agent is to maximize some, usually probabilistic, function of its cumulative Reward by means of modifying its Policy. The model is typically time discrete, stochastic, and usually assumed Markov, that is the next Environment State depends on the present State and Action only, and not in the further past. Formally stated, we have

$$G_t = R_{t+1} + \gamma R_{t+2} + \gamma^2 R_{t+3} + \ldots\ldots = R_{t+1} + \gamma G_{t+1} \qquad (1)$$

where $G_t$ is the cumulative Reward (usually called the Return), $R_{t+i}$ , i = 1,2,… are instantaneous Rewards, all referring to a time t, and $0 \leq \gamma \leq 1$ is a discount factor which accounts for the relative significance we assign to future Rewards (and may also guarantee convergence in case of infinite Reward series). The recursive form at the RHS of (1) is important as we will momentarily see. $R_{t+i}$, $S_t$, $A_t$ (as well as $G_t$) are all random variables governed by the Environment model distribution $p(s', r \mid s, a)$ (shorthand for $\Pr\{S_{t+1}= s', R_{t+1} = r \mid S_t = s, A_t = a\}$; we will apply similar abbreviations for other distributions), the Policy distribution $\pi(a \mid s)$ and the States distribution $\mu(s)$. The sum (1) may have a finite number of terms or not depending on whether the model is episodic or not. To assess how good a State is we may use its Value, which is defined as the mean of the Return assuming we start from this specified State and act according to a fixed Policy $\pi$

$$V(S_t = s) = E[G_t \mid S_t = s] = E[R_{t+1} + \gamma G_{t+1} \mid S_t = s] \qquad (2)$$

where E[ .] is the expectation operator, and where the right side equality is based on the recursive form of (1). For a non-specified State realization s, the Value $V(S_t)$ is a function of the r.v. $S_t$ and is thus a r.v. too. Assuming the model is indeed Markov we may derive from (2) to the important Bellman equation[3]

[3] The reader may refer to https://stats.stackexchange.com/questions/243384/deriving-bellmans-equation-in-reinforcement-learning for a neat proof of the Bellman equation by Jie Shi

$$V(S_t = s) = E[R_{t+1} + \gamma\, V(S_{t+1}) \mid S_t = s] \tag{3}$$

Many (but not all) of the methods used by RL are based on the Bellman equation. If the involved distributions p and $\pi$ are explicitly known then (3) may be directly evaluated. More commonly we may apply the law of large numbers, and iteratively approximate (3) by sampling $V(S_t)$ and $V(S_{t+1})$ and calculating

$$V(S_t = s)_{new} = V(S_t = s)_{old} + \alpha\, [R_{t+1} + \gamma\, V(S_{t+1} = s')_{old} - V(S_t = s)_{old}] \tag{4}$$

which is a slight generalization of the moving average formula, and where $\alpha$ is an update parameter. When (4) converges the quantity in square parenthesis (appropriately called 'the error') vanishes so that it may become a legitimate goal in a minimization problem. If the Value is approximated by some neural network parameterized by $W^V_t$, as is many times the case, we can iteratively update $W^V_t$ by means of stochastic gradient ascent

$$W^V_{t+1} = W^V_t + \eta^V\, [R_{t+1} + \gamma\, V(S_{t+1}; W^V_t) - V(S_t; W^V_t)]\, \nabla V(S_t; W^V_t) \tag{5}$$

where $\eta^V$ is the learning step parameter, where the gradient is calculated w.r.t. $W^V_t$ and where the factor within square parenthesis may be recognized as a parameterized version of the error in (4) above. Note that the gradient operator in (5) applies solely to $V(S_t; W^V_t)$ and not to $V(S_{t+1}; W^V_t)$ which also depends on $W^V_t$: this scheme constitutes a common and convenient formal abuse in RL methodology and is known as semi-gradient.

## 3. Learning Representations by means of Reinforcement Learning

Forcing an Agent to learn a task, for which success the ability to correctly discriminate between different classes of objects is required, should end up with the Agent possessing sufficiently rich internal representations of these classes. After the task training is completed the resultant representations (more precisely the mapping which generates them) can be attached to a linear classifier (e.g. single layer perceptron, SVM, etc.) and used as a feature extractor.

We demonstrate this idea by means of a simple example using the MNIST dataset. To make matters simpler we will define the Agent task itself as that of a classifier, that is it should predict which class (digit in our case) is being presented to it by the Environment (any other task requiring class discrimination will make it). The Agent will be rewarded by a small Reward each time it succeeds; by a bigger Reward after N consecutive successes, after which the episode ends and a new episode is launched; and penalized by a negative Reward upon each error.

RL comprises today a rich arsenal of methods by which we can approach this problem. We choose here the so called Policy Gradient scheme (Sutton et al, 2000; Sutton and Barto, 2018) where the Value, approximated by a neural network and the Policy, approximated by another, are concurrently trained so as to optimize a suitable goal. We define our goal here to be the maximization of the mean, over the state distribution $\mu(s)$, of the Value $V(S_t; W^V_t)$. This goal is translated into approximating the Value function by (5) above for all States, and learning the optimal Policy $\pi(a \mid s; W^\pi_t)$ by

$$W^\pi_{t+1} = W^\pi_t + \eta^\pi\, [R_{t+1} + \gamma\, V(S_{t+1}; W^V_t) - V(S_t; W^V_t)]\, \nabla \log \pi(A_t \mid S_t; W^\pi_t) \tag{6}$$

where $\eta^\pi$ is the Policy learning step parameter and where the gradient is taken w.r.t. the parameters $W^\pi_t$ of the Policy network (the derivation of (6) above is non-trivial and may be of interest, see e.g. Sutton and Barto, 2018 for details). Both neural nets are concurrently trained by backpropagation.

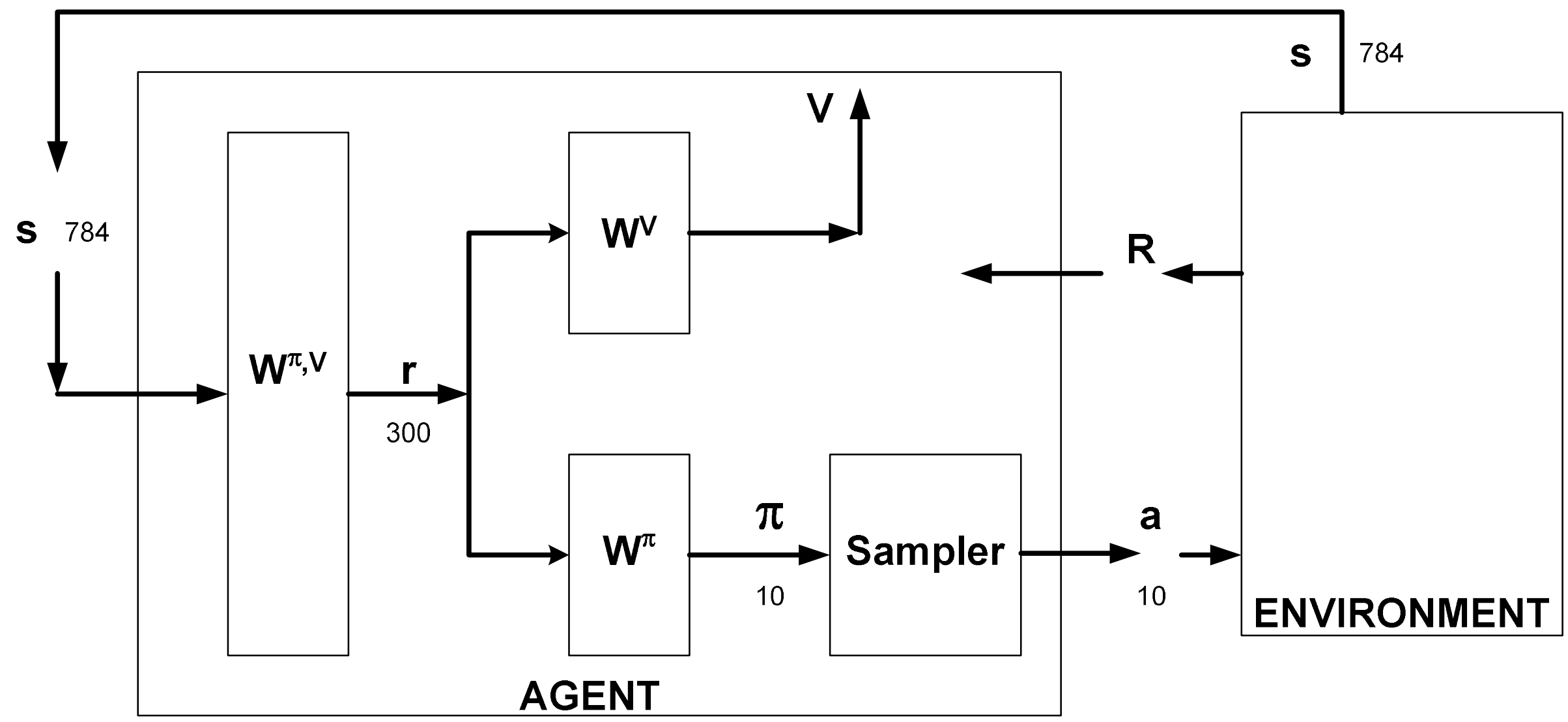

**Figure 1**: Simple demonstration of Representations Learning by means of Reinforcement Learning. All but last layers of $\mathbf{W}^{\pi}$ and $\mathbf{W}^{V}$ are shared and denoted as $\mathbf{W}^{\pi,V}$. The last $\mathbf{W}^{\pi}$ layer serves also as classifier layer, and **r** is the learned Representation.

In practice it is convenient to let all but the last layer of the neural nets to share the same parameters so we set [784 300] as dimensions for the shared layers ($784 = 28^2$ is the dimension of an MNIST digit vector) and 1 and 10 dimensions for the Value and Policy networks last layer respectively (all excluding a bias element). Please refer to Figure 1 above. We used Leaky ReLU activation functions for both nets hidden layers and for the Value net last layer, and softmax for the Policy net last layer. The softmax output of the Policy net, expressing the probability of choosing each of the possible Actions, was sampled to generate the class prediction. We assigned a non-Terminal Reward ( = +1) to each successful prediction, a Terminal Reward (= +10) after N (= 5) consecutive successes, and a negative Reward ( = -5) after each error.
We set $\eta^{\pi} = \eta^{V} = 1\text{e-}3$ and $\gamma = 0.9$[4]. We trained the system permutating the Training set every epoch until it achieved 0.4 % Error rate on the Training set (see Figure 2 below) after 880 epochs. We then measured the Classification Error rate with the MNIST Test set, with frozen parameters: 1.98 %. For reference purpose we trained with the same MNIST dataset a neural net classifier of same dimensionality ([784 300 10]) as our Policy network above: it achieved 0.4 % Training set Error rate after 42 epochs (annotated for reference; and 0 % Training set Error rate after 53 epochs), and a Test set Error rate of 2.09 %, measured after attaining 0% Training set Error rate, negligibly inferior to our RL scheme above.

Note that in our demonstrations above we are not attempting to beat any MNIST Test Error rate record; our goal here is to provide a simple comparative analysis between RL and a 'standard' neural net with similar function approximation capacity.

The learning steps in both demonstration setups were approximately set at the maximum values that lead to robust convergence. We observe in passing a large gap in learning rate (880 vs. 42 epochs till 0.4 % Error rate as stated above) between both classifiers; it is premature to state whether this reflects a fundamental difference between the schemes, or not; and this falls beyond the scope of this paper.

Before we proceed it may be of interest to qualitatively probe whether learning representations under a less demanding task may still provide full inter-class discrimination, or just the minimal required to fulfill the task. For this purpose we trained a RL Policy Gradient network similar to the above to solve the task of 'Picking' a presented MNIST digit if 'odd' and 'Passing' it if 'even'. After training to high accuracy w.r.t. this task (about 99%), the first layer of the network, mapping the learned representations (of dimension 300), was frozen, and adjoined to a linear classifier, which was trained and tested for digits ('0' to '9') Classification accuracy.

[4] Matlab code of this simple setup will be available upon request from the Author

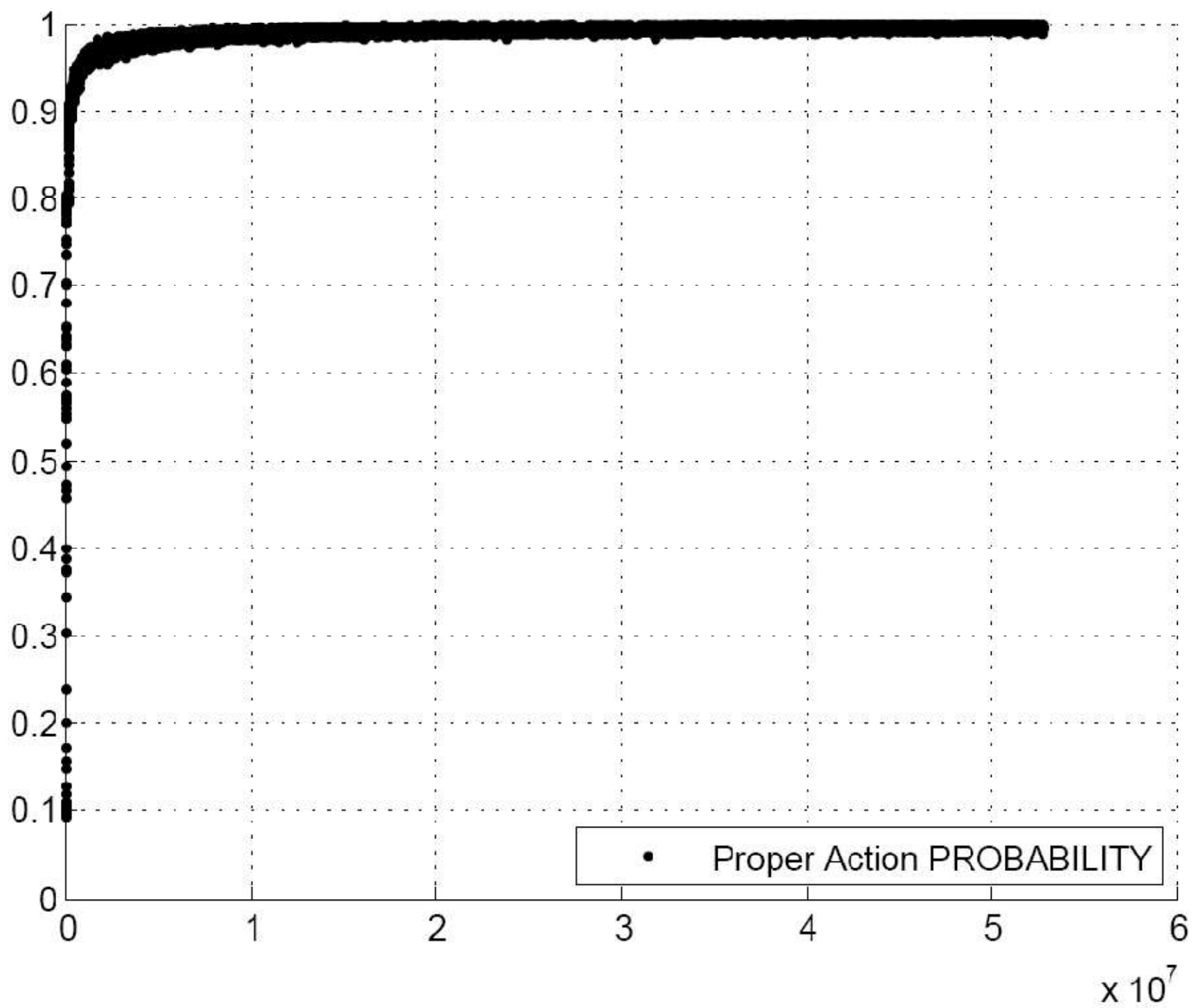

**Figure 2**: Training set Pr{Proper Action}, equivalent to (1 – Pr{Training Error}), starts with a brief slightly descending segment (unseen) due to errors penalty, followed by a rapid uprise, and ending in slow final convergence to 0.4 % Training set Error rate after 880 epochs, and still going.

As might be expected accuracy was poor (about 72%), reflecting the fact that the learned representations, during the RL training session, are able to almost perfectly discriminate between 'odd' and 'even' digits but, apparently, not much more. This failure may well also explain failures of the fore mentioned Unsupervised Learning "auxiliary tasks" methods (e.g. Donahue et al., 2017; etc.); these develop internal representations fit for the job at hand (e.g. puzzle solving, etc.) but not much more and certainly not for full inter-class discrimination.

We shall next see what are the circumstances under which RL may be considered a legitimate Unsupervised (or, alternatively, Supervised) Learning paradigm.

## 4. Unsupervised or Supervised Learning?

The following discussion might appear to some readers to be of philosophical nature (no offense implied), or alternatively, trivial. Nevertheless, it is important; and deeper than what it may appear.

It might be claimed by the astute reader that while the Agent in the demonstration of the last Section evidently does not need labels during its learning process (Rewards and States are its only inputs), the Environment does need. It has to have access to MNIST digits labels in order to 'decide' which type of Reward to assign (e.g. positive for correct prediction, negative for error, etc.). And so we may not consider the process as a whole, at system level, to be of Unsupervised Learning nature. This of course is true.

To crystallize we conduct 2 simple thought experiments:

Consider a *real world* Environment with a *real* Cat playing the Agent; by loose analogy we may call this an "in vivo" experiment. The *abstract entity* which we call Environment may present to the Cat a *real* Dog from which the Cat should escape to get a Reward (*real* safety in this case); or it may present it a *real* Mouse which the Cat should persecute in order to get a Reward (*real* fun in this case). Failing to act appropriately may result in a negative Reward (a *real* bite in case of the Dog or *real* boredom in case of the Mouse). After some reflection it should be easy to see that no decision from the part of the abstract Environment is required in order to assign a Reward: these are given, well, 'naturally'. And as a result, no label is involved in the process. The RL paradigm in this case should be considered a legitimate Unsupervised Learning scheme.

In sharp contrast to this we now consider a similar situation, but with a *simulated* Environment and a *simulated* Agent; by the same analogy this may now be called an "in vitro" scenario. The *real entity* which we call Environment (now a piece of computer code) may present to the Cat a *simulated* Dog (for example a photograph of a Dog) from which the Cat has to 'escape' in order to get a *simulated* Reward (such as a positive valued scalar), etc. The *simulated* 'escape' Action (assuming this is the selected Action) is presented to the *simulated* Environment in the form of some set of bits (say, or some other symbolic form). The Environment then, has to use the fore mentioned symbol representing the Action and the *label* of the presented State ('Dog'), which was provided by some genie, and use them combined to produce the appropriate Reward (e.g.by means of a look-up table). In the absence of a State *label* the process gets stuck[5]. We have no choice but to consider this RL case as a Supervised Learning scheme.

The 2 experiments above present extreme cases for the sake of clarity. In practice we may also envision scenarios where a *real* Agent acts upon a *simulated* Environment, or vice versa. The key point in any scenario is how the Reward assignment is produced by the Environment: whether 'naturally' (no labels required, Unsupervised Learning) or 'artificially' by means of some symbol processing (labels required, Supervised Learning).

## 5. Discussion and Conclusions

We have presented a comprehensive body of previous work. This has lead us to deduce that there exists today, after decades of research, a performance gap of about 20% between the recognition accuracy of our man-devised Unsupervised Learning methods and that of Unsupervised Learning as exhibited in the animal (specifically human) world. Animal Unsupervised Learning achievements naturally serve as our source of inspiration (just as flying birds inspired men to invent ways to fly), so that it was and is reasonable for us to expect that a competitive Unsupervised Learning scheme indeed exists; we just apparently happen to not have uncovered it yet. We proposed herein a possible resolution to this apparent puzzle.

As demonstrated above by a simple setup, Reinforcement Learning may force an Agent to learn rich enough representations to allow it to discriminate amongst classes as much as needed to successfully fulfill the task at hand. We have reasoned that such RL schemes, when implemented in a real world Environment should be considered to be of true Unsupervised Learning nature, slightly refining our conventional taxonomy (we may have been all along victims of our own terminology) which exclusively divides paradigms into *either* SL *or* UL *or* RL. RL boasts

[5] We could consider in principle the case wherein the Environment predicts by itself a label from the Dog's photograph; this however requires the Environment to be a trained classifier, which enters us into an endless loop

recognition competitive performance, as we explicitly illustrated in our MNIST Toy case above and as it was implicitly demonstrated by a myriad of RL experiments (e.g. Hessel et al. 2018). *Hence, RL may actually be that missing piece of our puzzle.*

As we indicated RL however, has the strange and unique peculiarity to metamorphose its nature when embodied in simulated Environments, and transform, with some exceptions, into a scheme of (not less) true Supervised Learning character.

This peculiarity might have been the source of our prolonged confusion. Indeed it is possible a result of this that RL has been usually considered in literature as a class of its own (i.e. neither Supervised nor Unsupervised) instigating further exploration efforts, to no avail so far.

Why all this matters? Because if RL turns out indeed to be the Unsupervised Learning scheme which the (real) animal world employs, then further search for the Holy Grail, a learning method with competitive accuracy which unsupervisedly learns representations in simulated environments, may be futile.

This is not to say that man-made systems of all kinds are always and inherently unable to competitively learn unsupervisedly. We may cite here 2 examples:

a. Robots acting in real world Environments (as might usually be the case). Such a Robot may e.g. learn by RL that when short of charge it should seek for a recharge station, its Reward being a real recharge. No labels are involved in this process.

b. Simulated Environments, but where the required labels are not provided by 'some genie', as mentioned above. In such cases the labels themselves are used in order to generate the associated objects; this may occur for example in systems (e.g. video games) where a label (e.g. 'Dog') is used to draw a dog caricature with some randomness in it, or to sample a realistic dog image from some distribution (e.g. by means of a generative network like GAN, etc). Labels in such cases are associated with their respective objects by design (not externally provided), and it seems reasonable to consider these cases to be of unsupervised learning nature as well.

Of course one or both of our conjectures above may turn out false: animal brains, our source of inspiration, may be unsupervisedly learning by a completely different scheme than RL which we may try, once discovered, to imitate; and/or there might exist ways of rich representations learning other than RL, which may remain unsupervised even in simulated environments. Time will tell: Reinforcement Learning in Animals, the Neuroscience of Reinforcement Learning, and Unsupervised Learning of Representations are all active areas of research.